\documentclass{article}

\usepackage{arxiv}

\usepackage[utf8]{inputenc} 
\usepackage[T1]{fontenc}    
\usepackage{hyperref}       
\usepackage{url}            
\usepackage{booktabs}       
\usepackage{amsfonts}       
\usepackage{nicefrac}       
\usepackage{microtype}      
\usepackage{lipsum}
\usepackage{graphicx}
\usepackage{amssymb, amsmath}
\usepackage{color}
\usepackage{url}
\usepackage{comment}
\usepackage{afterpage}
\usepackage{here}
\usepackage{mathrsfs}
\newcommand{\reffig}[1]{Figure~\ref{#1}}

\newcommand{\reftab}[1]{Table~\ref{#1}}

\newcounter{alnum}

\newcounter{rmnum}

\newcounter{muni}

\newcommand{\bi}[1]{\ensuremath{\boldsymbol{#1}}}

\title{Inverse airfoil design method for generating varieties of smooth airfoils using conditional WGAN-gp}

\author{
	Kazuo Yonekura \thanks{\texttt{yonekura@struct.t.u-tokyo.ac.jp}}\\
	Department of Systems Innovasion\\
	The University of Tokyo\\
	Tokyo, JAPAN 113-8656 \\
	\And
	Nozomu Miyamoto \\
	Department of Technology Management for Innovation\\
	The University of Tokyo\\
	Tokyo, JAPAN 113-8656 \\
	\AND
	Katsuyuki Suzuki \\
	Department of Systems Innovasion\\
	The University of Tokyo\\
	Tokyo, JAPAN 113-8656 \\
}

\begin{document}
	\maketitle
	
	\begin{abstract}
		Machine learning models are recently utilized for airfoil shape generation methods. 
		It is desired to obtain airfoil shapes that satisfies required lift coefficient. 
		Generative adversarial networks (GAN) output reasonable airfoil shapes. 
		However, shapes obtained from ordinal GAN models are not smooth, and they need smoothing before flow analysis.
		Therefore, the models need to be coupled with Bezier curves or other smoothing methods to obtain smooth shapes. 
		Generating shapes without any smoothing methods is challenging. 
		In this study, we employed conditional Wasserstein GAN with gradient penalty (CWGAN-GP) to generate airfoil shapes, and the obtained shapes are as smooth as those obtained using smoothing methods. 
		With the proposed method, no additional smoothing method is needed to generate airfoils.
		Moreover, the proposed model outputs shapes that satisfy the lift coefficient requirements. 
	\end{abstract}


\section{Introduction}
\label{intro}
Airfoil designs incur intensive computation costs. The airfoil design task is to obtain airfoil shapes that satisfy specific requirements. 
Recently, shapes have been designed iteratively using computational fluid dynamics (CFD). CFD-based design optimization methods have been studied for years \cite{Jameson95,Shelton93}. 
However, CFD computation is a time-consuming process. 
Recently, data-driven methods, such as machine learning, have been employed for airfoil generation. 
Once a machine learning model is trained, it outputs immediately. As the model can be reused for other design tasks, the computation time is sufficiently smaller than that of CFD-based iterative methods. 
Several studies on generative adversarial networks (GAN) have been reported.
\cite{wang21} proposed airfoil GAN that generates airfoil shapes regardless of requirements. 
Airfoil generally needs smooth curves, but output shapes using GAN are not always smooth due to noise during learning and generation. 
To obtain smooth shapes, \cite{Chen21} proposed {B}{\'e}zier{GAN} in which airfoil is represented by B{\'e}zier curves, which successfully produced smooth curves. 
\cite{Achour20} generated low- and high-lift-coefficient airfoils using conditional GAN (cGAN). 
Most shapes reported in the literature have jaggy lines, which hinder further CFD analyses. 
Consequently, the Savitzky--Golay filter \cite{Press90} has been employed to smoothen the curves. 
To date, generating smooth airfoil shapes without using a smoothing method or free curves is challenging.
Herein, we propose conditional Wasserstein GAN with gradient penalty (cWGAN-gp) model to overcome this issue. 
Furthermore, in existing cGAN models \cite{Achour20}, the conditional label is binary, i.e., high or low, and it cannot generate airfoils that have specific lift coefficients. 
The proposed model uses the lift coefficient as a continuous label; therefore, shapes that have a specific lift coefficient are obtained.

Another method utilizes variational autoencoder (VAE) models for airfoil generation.
\cite{Yonekura21a} used conditional VAE to generate airfoils that meet specific lift coefficients. \cite{Yonekura21b} compared two types of conditional VAE (CVAE) model, the CVAE with normal distribution and von Mises Fisher distribution, and investigated the generated shapes. They also proved that if different types of airfoils are combined and fed as training data, the CVAE model generates airfoils that have mixed features of the different airfoils used. 
Shapes obtained from these VAE models are smooth enough for CFD analysis.
However, whether smooth airfoils can be generated using GAN is still an open problem.
One of the differences between GAN and VAE is that the loss function in VAE is generally element-wise metrics between input and generated data, and VAE does not consider the properties of the data. 
GAN is useful because it considers not the element-wise metrics but the properties of the airfoils. 
VAEGAN \cite{VAEGAN} coupled VAE and GAN to combine the features of VAE and GAN.

Besides airfoils, data generation methods for various applications, including marine propeller \cite{Gaggero19} and turbine film-cooling holes, as well as airfoils \cite{YW14} and architecture \cite{Brown19}, have been reported. 
\cite{Oh19} used a deep generative model with topology-optimized shapes to develop various types of optimal shapes. 
Generative models are useful for three-dimensional (3D) objects \cite{bidgoli19}. 
\cite{Nash17} employed generative models for 3D shapes and generated 3D aircraft shapes. However, flow analysis was not conducted on the generated shapes. 
Employing the shape generation methods in various fields is easy if smoothing methods are not required.

Various GAN models have been proposed previously. 
Ordinal GAN uses the Jensen--Shannon (JS) divergence to measure the similarity of probability distributions. 
However, instability in the training of GAN, such as mode collapse \cite{goodfellow2017nips} and gradient dissipation \cite{Arjovsky17a}, has been reported 
Mode collapse is a situation wherein a generator outputs the same data even when the input latent vector is different.
Wasserstein GAN (WGAN) was proposed to overcome gradient dissipation \cite{Arjovsky17b}. WGAN uses the earth mover's (EM) distance to measure the distance of probability distributions. 
WGAN with gradient penalty (WGAN-gp) \cite{Gulrajani} was proposed to improve stability in training WGAN. 
To achieve our goal, we propose conditional WGAN-gp, which uses WGAN-gp conditionally. Conditional GAN usually employs discrete labels, but because we aim to obtain continuous lift coefficients, the continuous label is employed. 

The rest of the paper is organized as follows: the dataset is described in section 2; section 3 introduces the cGAN and WGAN-gp models; the introduced GAN models were trained using NACA airfoil data, as described in section 4, and the results are compared; the conclusion of the study is presented in section 5. 

\section{Airfoil data}
\begin{figure}[tbph]
	\begin{center}
		\begin{minipage}[h]{0.5\textwidth}
			\begin{center}
				\includegraphics[width=0.8\textwidth]{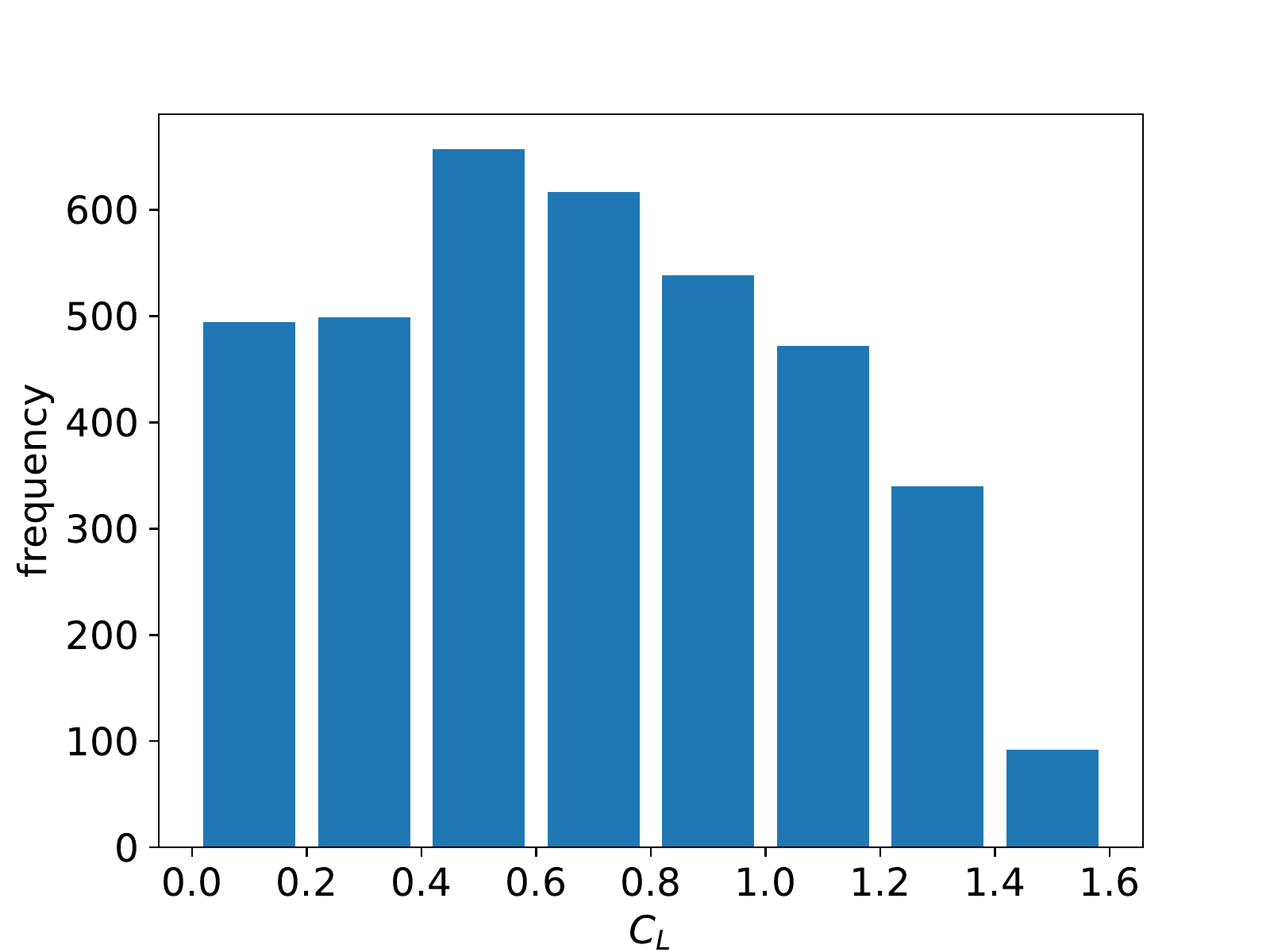}
			\end{center}
		\end{minipage}%
		\caption{Histogram of $C_L$. }
		\label{fig:hist}
	\end{center}
\end{figure}

NACA 4-digit airfoil data \cite{Abbot} were used for the training. 
NACA 4-digit airfoil is defined by three parameters: max camber, the position of the max camber, and thickness. 
One thousand data of four-digit airfoils can be defined, i.e., from 0000 to 9999, but some of them are not useful for airfoils. Herein, the lift coefficient was calculated for all the four-digit airfoils, and those whose $C_L$ could not be calculated or $C_L<0$ or $C_L>2.0$ were eliminated. 
After elimination, the total number of airfoils was 3709. 
The calculation condition of $C_L$ was as follows: angle of attack was 5 degrees, and Reynolds number was $3.0 \times 10^6$. 
Xfoil \cite{XFOIL} was used for calculating $C_L$, which employed a panel method for computation. 
The histogram of $C_L$ is shown in Fig. \ref{fig:hist}. The data are almost uniform, except that the amount of data for $1.2 < C_L$ is relatively small. 

\begin{figure}[tbph]
	\begin{center}
		\begin{minipage}[h]{0.5\textwidth}
			\begin{center}
				\includegraphics[width=\textwidth]{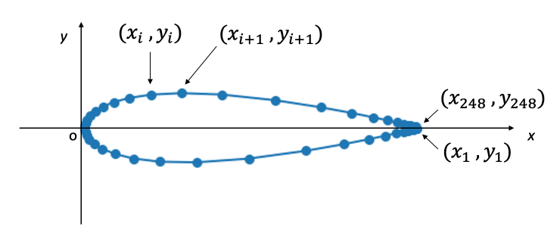}
			\end{center}
		\end{minipage}%
		\caption{Example of airfoil discretization. }
		\label{fig:disc}
	\end{center}
\end{figure}

To handle airfoils, 248 points on the airfoil outline were sampled. 
Coordinates $x_i, y_i$ were gathered to form a vector $\bi{x} = \left( x_1, \dots, x_{248}, y_1, \dots, y_{248} \right)$. An example of airfoil discretization is shown in Fig. \ref{fig:disc}.
The Xfoil computation was conducted using the 248 data points.
Xfoil needed more than $120$ points around the airfoil for stable computation. 

\section{GAN and VAE models}
\subsection{GAN and cGAN}
\begin{figure}[h]
	\begin{center}
		\begin{minipage}[h]{0.5\textwidth}
			\begin{center}
				\includegraphics[height=50mm]{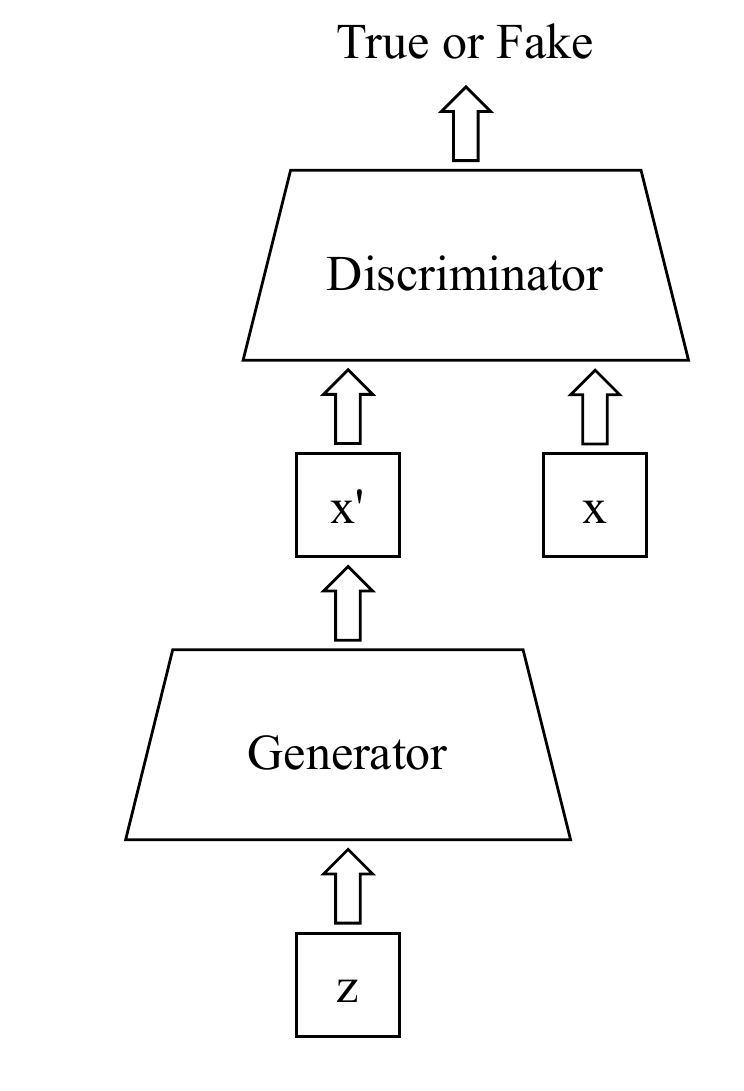}
				\par
				{(a) GAN}
			\end{center}
		\end{minipage}%
		\begin{minipage}[h]{0.5\textwidth}
			\begin{center}
				\includegraphics[height=50mm]{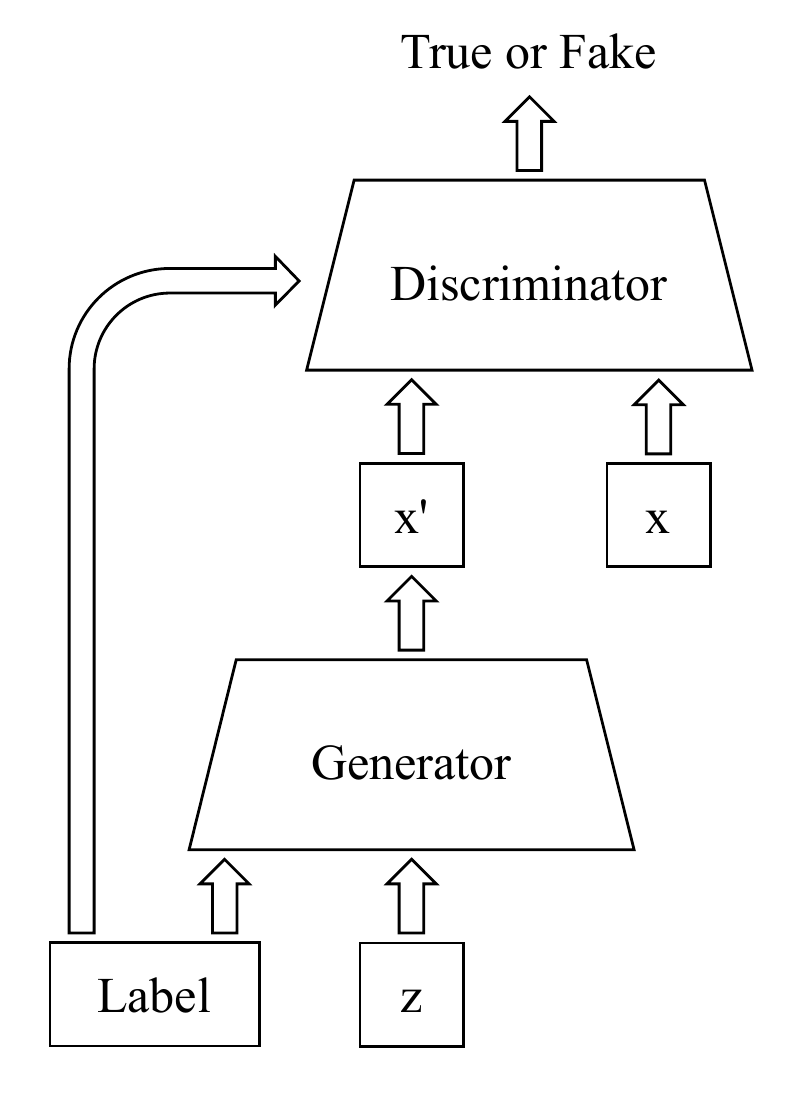}
				\par
				{(b) Conditional GAN}
			\end{center}
		\end{minipage}%
		\caption{Architecture of GAN and cGAN.}
		\label{fig:GAN}
	\end{center}
\end{figure}

GAN model comprises a generator and a discriminator. 
The generator inputs latent vectors $\bi{z}$ and outputs shape vectors $\bi{x}$.
The discriminator inputs shape vectors $\bi{x}$ and outputs whether the input data is true or fake. 
The architecture of the network is illustrated in \reffig{fig:GAN} (a). 
The generator and discriminator are represented by $G: \bi{z} \mapsto \bi{x}$ and $D: \bi{x} \mapsto [0,1]$, respectively, where $D(\bi{x})=1$ implies that discriminator judged the input $\bi{x}$ to be true data, and $D(\bi{x})=0$ implies that the data were fake.
GAN model was trained based on $\min_{G} \max_{D} V(D, G)$. The loss function $V(D, G)$ is formulated as
\begin{align*}
	V(D, G) &= \mathbb{E}_{\bi{x} \sim p_r(\bi{x})} [ \log{D(\bi{x})} ]
	+ \mathbb{E}_{\bi{z} \sim p_z(\bi{z})} [\log{ (1-D(G(\bi{z})) ) } ] , \\
	&= \mathbb{E}_{\bi{x} \sim p_r(\bi{x})} [ \log{D(\bi{x})} ]
	+ \mathbb{E}_{\bi{x} \sim p_g(\bi{x})} [\log{ (1-D(\bi{x}) ) } ], 
\end{align*}
where $E[\cdot]$ implies expectation, and $p_r(\bi{x})$ $p_z(\bi{x})$ represents the probability of occurrence of $\bi{x}$ and $\bi{z}$.
The network was optimized to minimize $V$ with respect to the generator and maximize $V$ with respect to the discriminator. 
The function $V$ indicates a higher value if the discriminator succeeds in determining whether the input data is true or fake. Otherwise, if the generator succeeds to cheat the discriminator, $V$ indicates a lower value because $1-D(G(\bi{z}))$ becomes smaller than 1. 
The generator was trained to cheat the discriminator, and the discriminator was trained not to be cheated by the generator. Because two networks were trained adversarial, the model is called a generative adversarial network. 

When the optimal $D=D^*$ is given, the loss function is expressed as 
\begin{align}
	V(D^*, G) = 2 D_{JS} \left( p_r || p_g \right) -2 \log 2, \label{eq.JS}
\end{align}
where $D_{JS}$ represents the JS divergence. 
JS divergence measures the distance between two probability distributions, i.e., the distribution of real and fake data. Thus, minimizing JS divergence implies matching $p_r$ and $p_g$.

Conditional GAN is developed to generate data with specific labels. 
The architecture of the cGAN is illustrated in \reffig{fig:GAN} (b). 
In the cGAN, both generator and discriminator has label $y$ to be input. 
Therefore, the generator outputs data with a specified conditional label. 
The discriminator judges fake or true considering the input data and label. 
The label $y$ is usually discrete, but herein, the continuous label was used to handle the continuous lift-coefficient label. 
The loss function is expressed as 
\begin{align*}
	V(D, G) = \mathbb{E}_{\bi{x} \sim p_r(\bi{x})} [\log{D(\bi{x}|y)} ]
	+ \mathbb{E}_{\bi{z} \sim p_z(\bi{z})} [\log{ (1-D(G(\bi{z}|y)) ) }]. 
\end{align*}

The generator sometimes outputs the same data $\bi{x}$ even if the input latent vector $\bi{z}$ is different. Thus, the function $G$ becomes an identity function. 
This situation is not desired, and it is called a mode collapse.

\subsection{WGAN-gp}
WGAN \cite{Arjovsky17b} has been proposed to overcome the gradient dissipation. 
The gradient dissipation in GAN is attributed to increased JS divergence in \reffig{eq.JS}.
In WGAN, Wasserstein distance, which is also called the EM distance, is employed. 
The EM distance $ W(\cdot, \cdot) $ between two probability distribution $p_r$ and $p_g$ is defined by 
\begin{align*}
	W(p _ r, p _ g) = \inf _ {\gamma \sim \Pi(p _ r, p _ g)} \mathbb{E} _ {(\bi{x},\bi{y}) \sim \gamma} \left[ \| \bi{x}-\bi{y} \| \right]. 
\end{align*}
Using the Kantorovich--Rubinstein duality \cite{Villani}, the EM distance is equivalent to 
\begin{align*}
	W(p _ r, p _ g) = \frac{1}{K} \sup _ { \|f\| _ {L} \leq K} \mathbb{E} _ { \bi{x} \sim p _ {r}} [f( \bi{x} )] - \mathbb{E} _ { \bi{y} \sim p _ {g}} [f( \bi{y})] ,
\end{align*}
where the supremum is taken over all $K$-Lipshitz functions. 
When a parameterized family of $K$-Lipshitz function $\{ f_w \} _{w \in \mathcal{W}}$ is given, the EM distance is calculated by 
\begin{align*}
	W(p_{r}, p_{g}) = \max _ {w \in \mathcal{W}} \left[  \mathbb{E} _ { \bi{x} \sim p _ {r}} [f _ {w}( \bi{x})] - \mathbb{E} _ { \bi{z} \sim p _ {z}} [f _ {w}(g _ {\theta}( \bi{z}))] \right] .
\end{align*}
In the neural network model, $f_w$ and $g_\theta$ correspond to the discriminator and generator, respectively. The parameter $w$ of the family $\{ f_w \}$ is the weights of the neural network. To guarantee $K$-Lipshitzness, weight clipping is employed. 
Using the EM distance, the loss function of WGAN is expressed as $V_{\rm WGAN}( f_w, g_\theta) = W(p_{r}, p_{g})$, and WGAN solves $ \min_\theta \max_w [ V_{\rm WGAN}(f_w, g_\theta) ]$. 

It is proven in \cite{Gulrajani} that the gradient of the optimal $f_w$ has norm 1 almost everywhere. Therefore, WGAN-gp adds a penalty term to ensure that the gradient of $f_w$ becomes 1. Consequently, the loss function of WGAN-gp is 
\begin{align*}
	V_{\rm WGAN\mathchar`-gp}(f_w, g_\theta) = V_{\rm WGAN} (f_w, g_\theta) + \lambda \mathbb{E} _ {x \sim p _ {r}} 
	[ \left( \| \nabla_{\bi{x}} f _ {w}(\bi{x}) \|_2 -1 \right) ^2] .
\end{align*}

\subsection{cWGAN-gp for airfoil generation}
The architecture of the cGAN was used with Wasserstein distance and gradient penalty. 
The label node input the lift coefficient as a continuous label. 
The dimension of latent $\bi{z}$ was set as $6$, and the architecture of the generator comprised five layers whose nodes were 64, 128, 256, 512, and 496.
The architecture of the discriminator comprised three layers whose nodes were 512, 256, and 1.
The hyperparameters in training are shown in \reftab{tab:param}.
The code was implemented using PyTorch \cite{paszke2017automatic}. 
The computation was performed on AMD EPYC 7402P 2.80GHz CPU with NVIDIA RTX A6000 GPU.

\begin{table}[htpb]
	\begin{center}
		\caption{Hyperparameters in training GAN models.}
		\label{tab:param}       
		\begin{tabular}{l|c}
			\hline\noalign{\smallskip}
			params & value \\
			\hline\noalign{\smallskip}
			optimization algorithm & Adam \\
			learning rate & 0.0001 \\
			number of training steps for D per iter. & 5 \\
			\noalign{\smallskip}\hline
		\end{tabular}%
	\end{center}%
\end{table}%

\begin{figure*}[tbph]
	\begin{center}
		\begin{minipage}[h]{\textwidth}
			\begin{center}
				\includegraphics[height=\textwidth,angle=90]{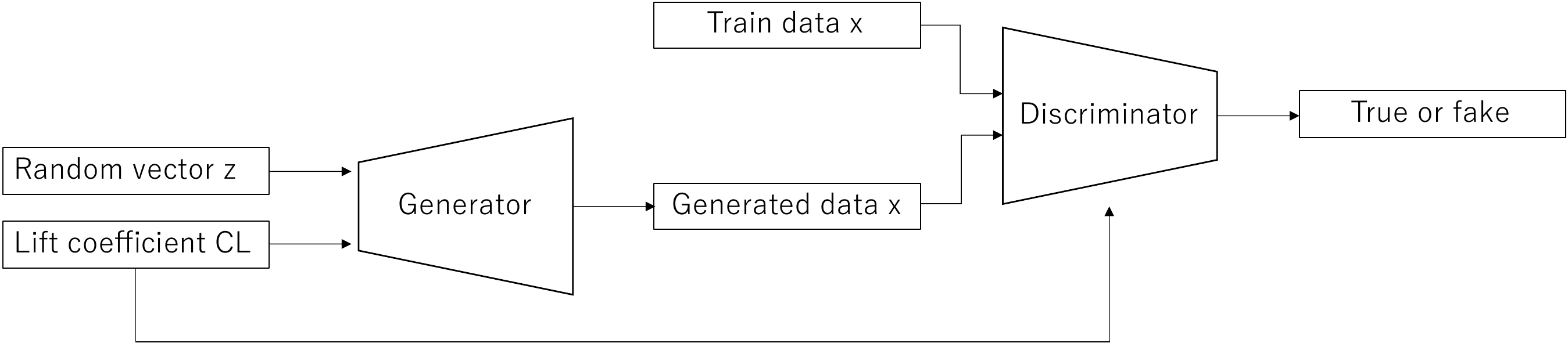}
			\end{center}
		\end{minipage}%
		\caption{GAN model for airfoil generation. }
		\label{fig:cWGAN}
	\end{center}
\end{figure*}

\subsection{$\mathcal{N}$-CVAE and $\mathcal{S}$-CVAE}
\begin{figure}[h]
	\begin{center}
		\begin{minipage}[h]{0.5\textwidth}
			\begin{center}
				\includegraphics[width=\linewidth]{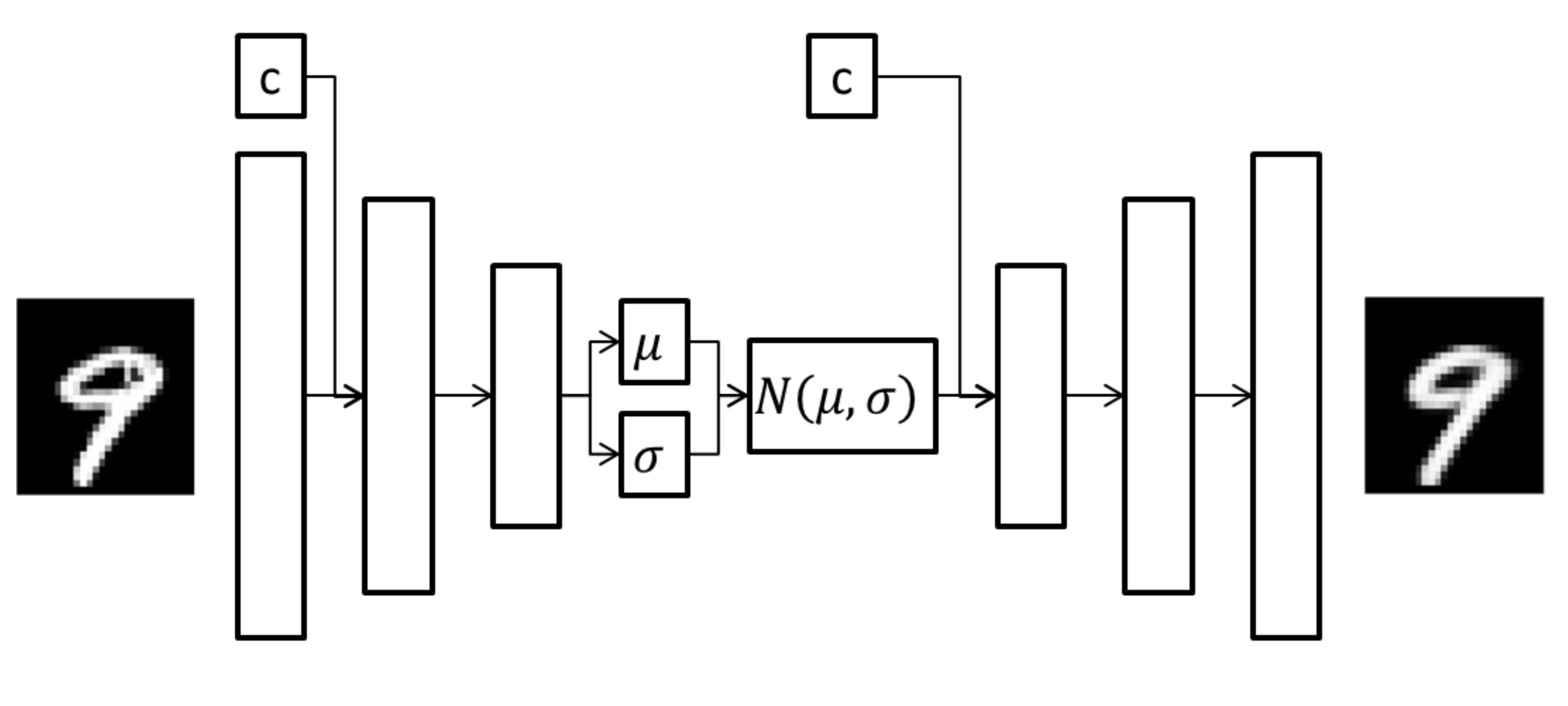}
			\end{center}
		\end{minipage}%
		\caption{Schematics of $\mathcal{N}$-CVAE.}
		\label{fig:CVAE}
	\end{center}
\end{figure}

The airfoil generation using $\mathcal{N}$-CVAE and $\mathcal{S}$-CVAE is studied in \cite{Yonekura21b}. We compare the results of GAN models with $\mathcal{N}$-{} and $\mathcal{S}$- CVAE models in the numerical experiments.
The CVAE model is one of deep generative models. 
The schematics of the $\mathcal{N}$-CVAE model is illustrated in \reffig{fig:CVAE}. The model consists of an encoder and a decoder. 
The encoder inputs data and label, and embeds features on a latent space. 
In $\mathcal{N}$-CVAE, the latent space is $d$ dimensional space, and the CVAE model is trained so that the data distribution on the latent space becomes standard normal distribution. 
The dimension $d$ is generally smaller than that of the input data, and the literature \cite{Yonekura21b} uses $d=2$. 
Both $\mathcal{N}$-{} and $\mathcal{S}$-CVAE models are trained to minimize the loss function $ \mathcal{L}_{CVAE}$ which is defined by
\begin{align}
	\mathcal{L}_{CVAE} 
	&= \sum \| \bi{x} - \bi{x}' \| + \mathcal{L}_{\rm KL}(p||q), \label{eq.VAE}
\end{align}
where $\bi{x}$ is the input data and $\bi{x}'$ is the generated data output from the decoder. 
$\mathcal{L}_{\rm KL}(p||q)$ implies the Kullback-Leibler divergence between the prior $p$ and the posterior $q$. 
In the $\mathcal{N}$-CVAE, the standard normal distribution is used as a prior, whereas in the $\mathcal{S}$-CVAE model, von Mises-Fisher distribution is used. 
A large difference between GAN models and VAE models is that the loss function of VAE contains distance between input and output data, whereas that of GAN does not. 

\section{Numerical example}
\subsection{Comparison with cGAN, cWGAN-gp, and other generative models}
cGAN and cWGAN-gp models were trained individually using the NACA dataset. The latent dimension was set as $d=3$ and $d=6$. 
Once trained, shapes were generated using labels in $C_L = 0.1, 0.11, \dots 1.5$, and 20 shapes were generated for each $C_L$ using random latent vector. 
Then, $C_L$ was recalculated using XFoil for the generated shapes. 
If the absolute error between the calculated and specified label was larger than 0.2, the shape generation was considered a failure. 
Among all the generated shapes, nonconvergence, success, and failure rates are shown in Tab. \ref{tab:success}. 
To measure the variety of the generated shapes, $\mu$ was calculated using the following equation;
$
\mu = \frac{1}{N} \sum_i \left \| \bi{x}_i - \sum_j  \bi{x}_j \right \| ^2. 
$
Comparing cGAN and cWGAN-gp, the nonconvergence rate was smaller in cWGAN-gp ($d=3$), which implied that the generated shapes were smoother. 
Moreover, the success rate was significantly large in cWGAN-gp ($d=3$). 

cGAN with smoothing method \cite{Achour20} is also compared in Tab. \ref{tab:success}. The data was adapted from the literature \cite{Achour20}, and the training dataset was different from that of cGAN and cWGAN-gp. Further, in \cite{Achour20}, the label is high or low lift by drag ratio, which is different from that obtained herein. Therefore, we could not directly compare the results to the labels. Therefore, the success and failure rates are shown in parentheses. 
The ordinal conditional VAE ($\mathcal{N}$-CVAE) and hyperspherical CVAE ($\mathcal{S}$-CVAE) models \cite{Yonekura21a,Yonekura21b} were also compared. 
The $\mathcal{N}$-CVAE model use standard normal distribution as a prior, and $\mathcal{S}$-CVAE uses von Mises-Fisher distribution. 
For these models, the training dataset and success threshold are the same as those obtained herein.
Although cWGAN-gp showed the highest nonconvergence ratio, no significant difference existed between cWGAN-gp, cGAN with smoothing, and $\mathcal{S}$-CVAE. Consequently, cWGAN-gp generated smooth shapes without any smoothing methods. 

The mean squared error (MSE) was smaller in cGAN than cWGAN-gp. 
Moreover, $\mu$ was smaller in cGAN than cWGAN. 
This difference indicates that the generated shapes from cGAN were similar. 
In cGAN, the success rate was small, and the generated shapes were in good agreement with the labels, although the variety of shapes was limited. 
Conversely, in cWGAN-gp, the success rate and variation of shapes were high, and error with respect to $C_L$ was larger than that of cGAN. 

Among all generative models, cWGAN-gp and $\mathcal{S}$-CVAE showed high $\mu$, and MSE of cGAN and cWGAN-gp was higher than that of $\mathcal{N}$-{} and $\mathcal{S}$-CVAE. 
Consequently, cWGAN-gp is the best among the GAN models. 
VAE models are better than GAN models considering MSE and $\mu$.
VAE is trained to minimize the distance between input and generated shapes, which is formulated in the first term of \reffig{eq.VAE}.
This term calculates point-wise errors of input and generated shape. 
Conversely, GAN models do not consider point-wise errors but the properties of the shapes, i.e., the generator is trained to cheat the discriminator. Therefore, in principle, VAE models cannot generate completely new shapes.

\begin{table}[tpb]
	\footnotesize
	\begin{center}
		\caption{Success rates and errors of generated shapes.}
		\label{tab:success}       
		\begin{tabular}{lrlrrrr}
			\hline\noalign{\smallskip}
			& not converge$\downarrow$ & & failure$\downarrow$ & success$\uparrow$ & MSE$\downarrow$ & $\mu$$\uparrow$  \\
			\noalign{\smallskip}\hline\noalign{\smallskip}
			cGAN $(d=3)$& 26.8\% & & 24.4\% & 48.8\% & 0.0470 & 0.152\\
			cGAN $(d=6)$ & 26.6\%& & 25.7\% & 47.7\% & 0.050 &0.139\\
			cGAN with smoothing \cite{Achour20} & 10\% & & (15\%) & (75\%) & - & - \\
			cWGAN-gp $(d=3)$ & $\boldsymbol{9.6}\%$ & & 15.3\% &  $\boldsymbol{75.1}\%$ & 0.0469 & $\boldsymbol{0.320}$ \\
			cWGAN-gp $(d=6)$ & 14.6\% & & 23.8\% &  61.6\% & 0.0374 & 0.319 \\
			\noalign{\smallskip}\hline\noalign{\smallskip}
			$\mathcal{N}$-CVAE \cite{Yonekura21a} & 12.0\% & & $\boldsymbol{13.0\%}$ & 75.0\% & 0.027 & 0.226\\
			$\mathcal{S}$-CVAE \cite{Yonekura21b} & 10.2\% & & 15.0\% & 74.8\% & $\boldsymbol{0.020}$ & 0.317\\
			\noalign{\smallskip}\hline& 
		\end{tabular}%
	\end{center}%
\end{table}%

\subsection{Generated shapes}
\begin{figure*}[tbph]
	\begin{center}
		\begin{minipage}[h]{0.5\textwidth}
			\begin{center}
				\includegraphics[width=\textwidth]{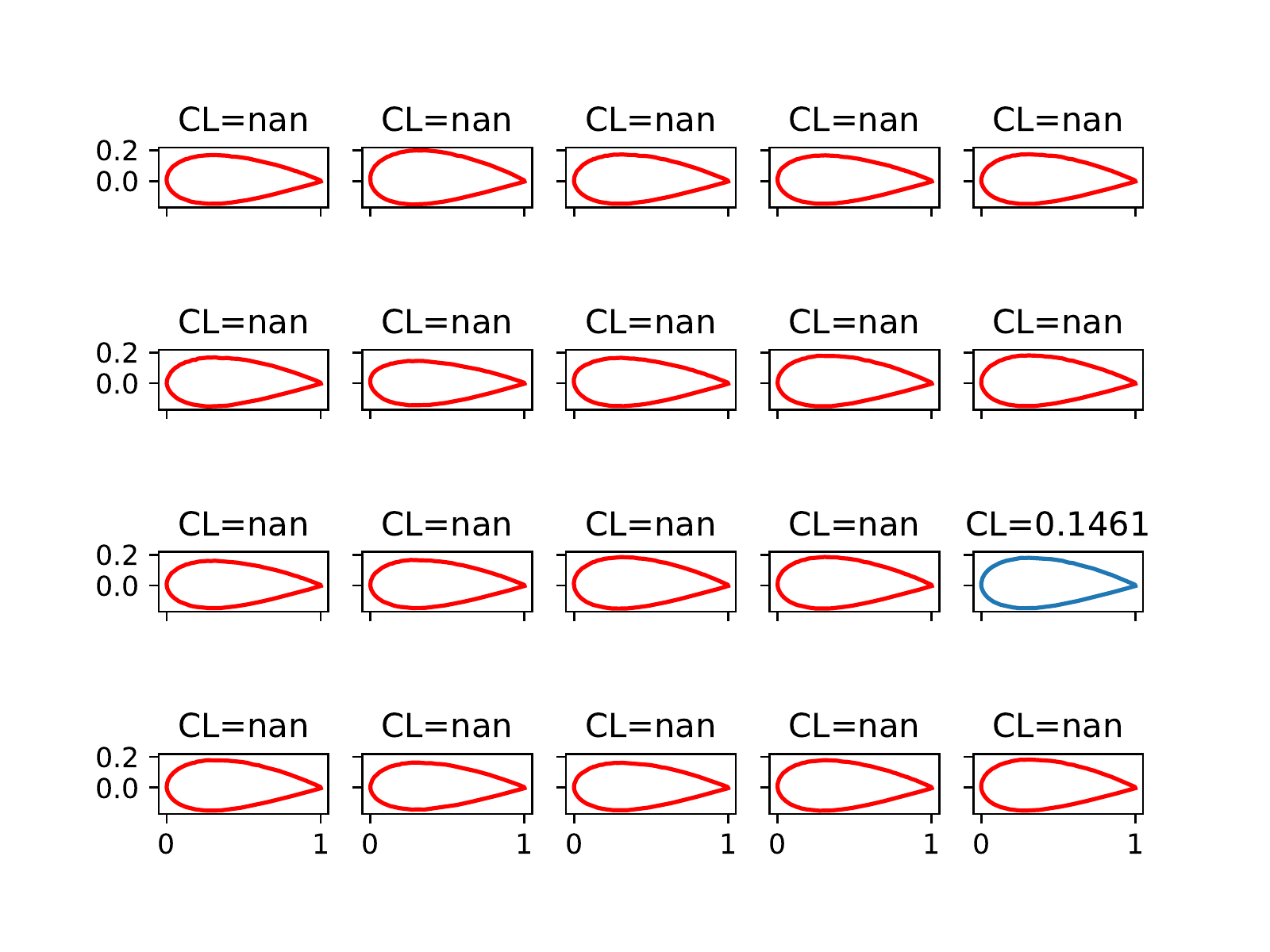}
				\par
				{\footnotesize (a) cGAN, $C_L=0.1$. }
			\end{center}
		\end{minipage}%
		\begin{minipage}[h]{0.5\textwidth}
			\begin{center}
				\includegraphics[width=\textwidth]{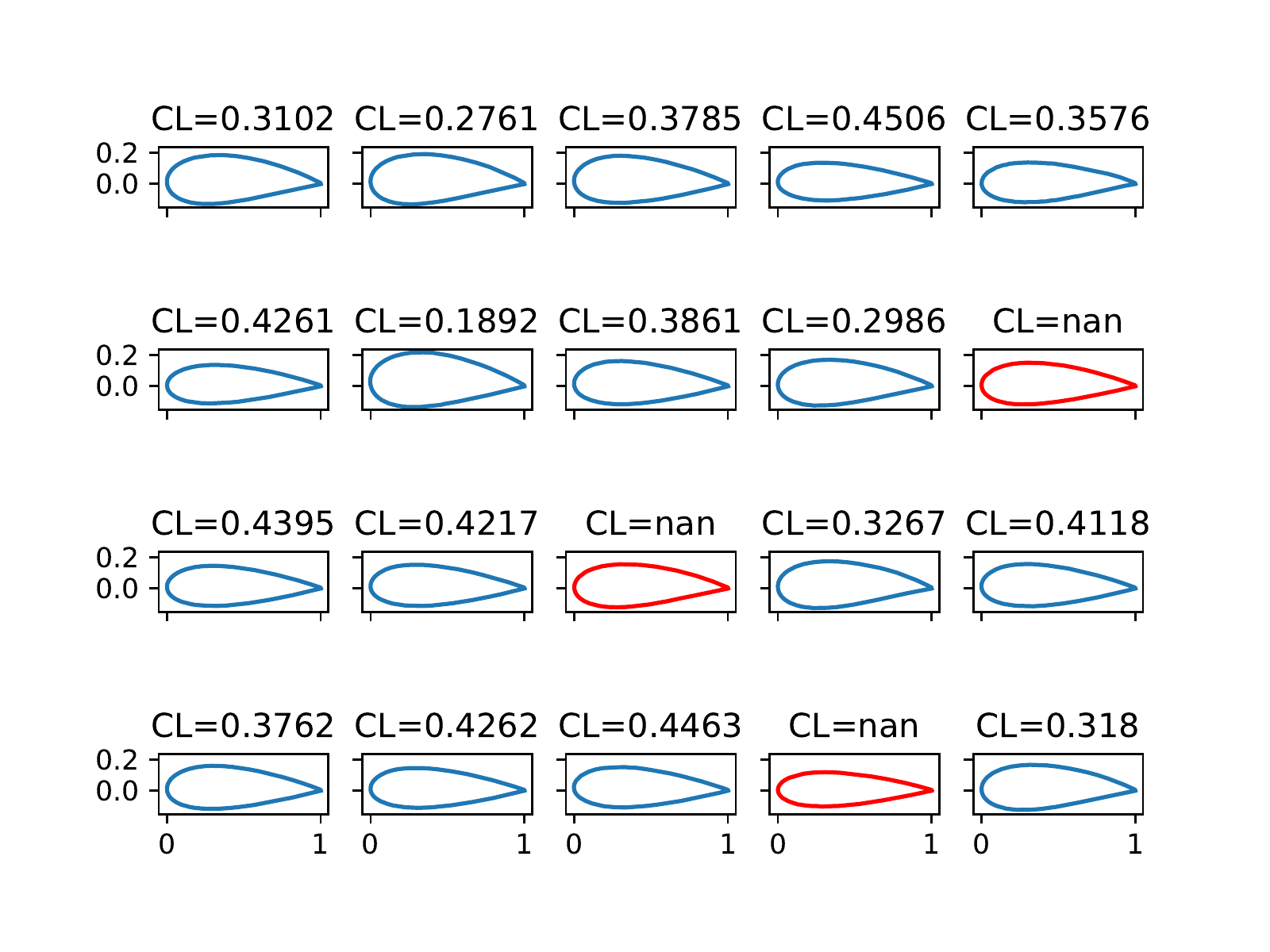}
				\par
				{\footnotesize (b) cGAN, $C_L=0.5$. }
			\end{center}
		\end{minipage}%
		\par
		\begin{minipage}[h]{0.5\textwidth}
			\begin{center}
				\includegraphics[width=\textwidth]{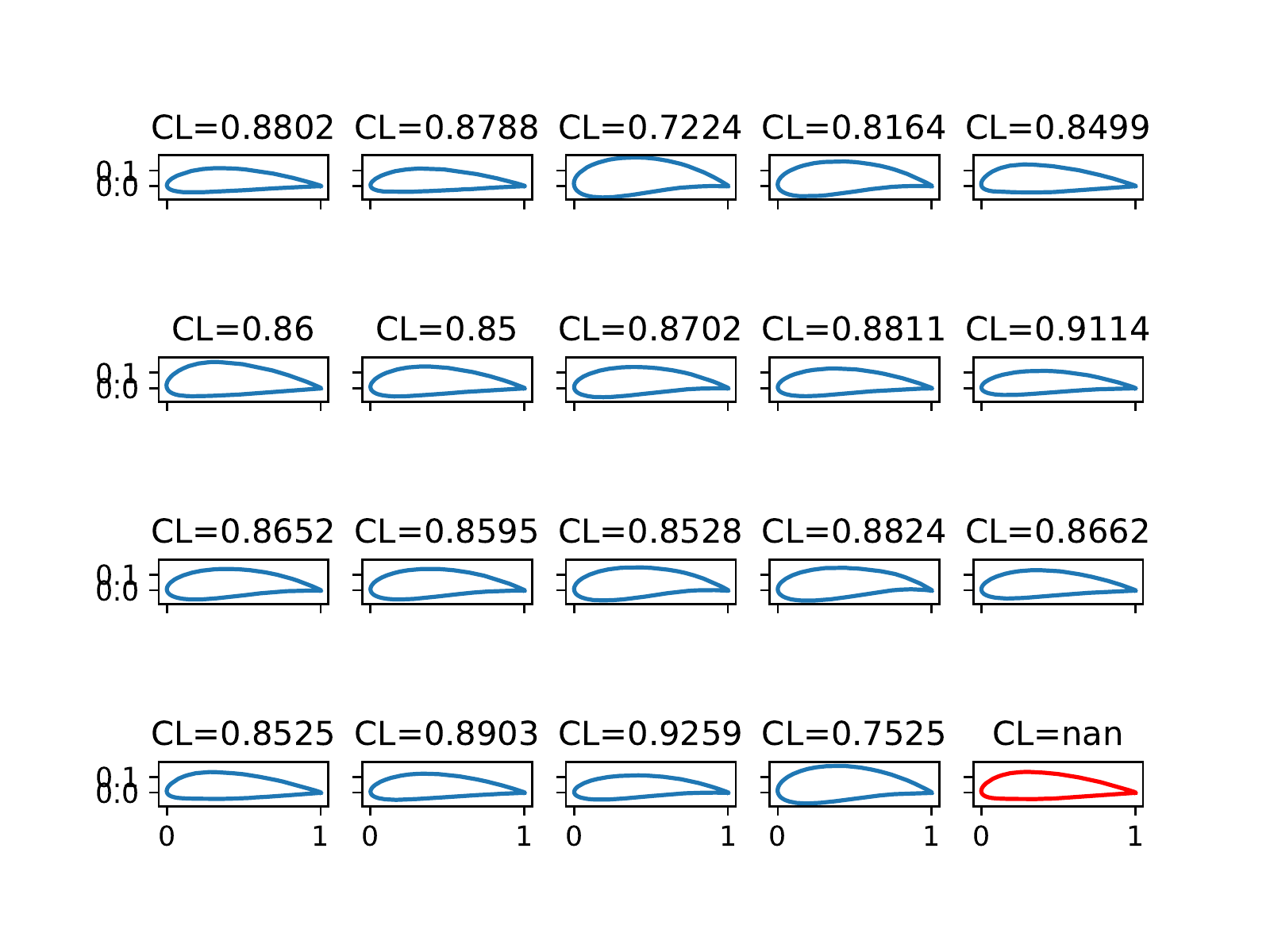}
				\par
				{\footnotesize (c) cGAN, $C_L=1.0$. }
			\end{center}
		\end{minipage}%
		\begin{minipage}[h]{0.5\textwidth}
			\begin{center}
				\includegraphics[width=\textwidth]{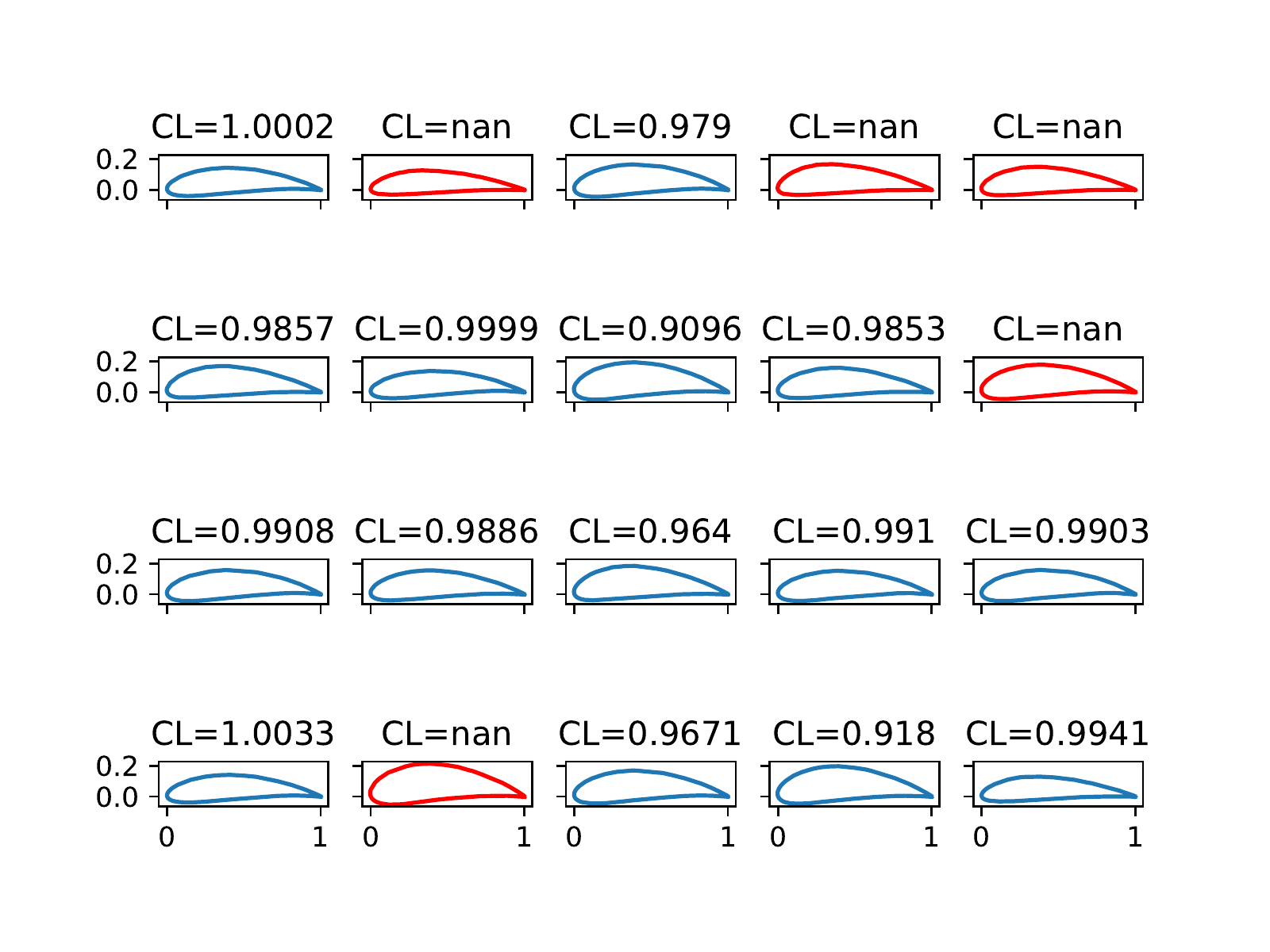}
				\par
				{\footnotesize (d) cGAN, $C_L=1.4$. }
			\end{center}
		\end{minipage}%
		\caption{Generated shapes of cGAN ($d=3$). }
		\label{fig:shapes_cGAN}
	\end{center}		
\end{figure*}

\begin{figure*}[tbph]
	\begin{center}
		\begin{minipage}[h]{0.5\textwidth}
			\begin{center}
				\includegraphics[width=\textwidth]{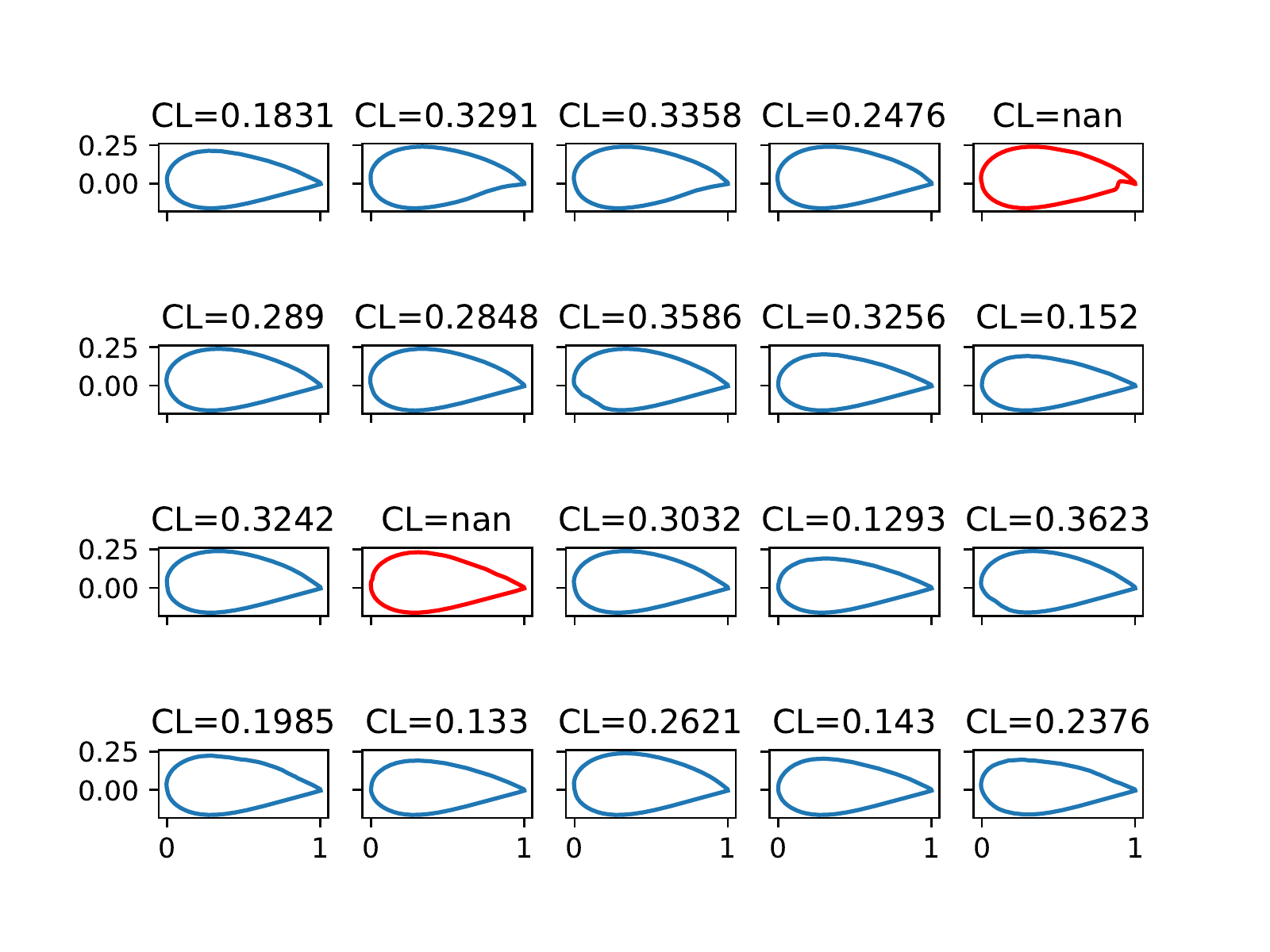}
				\par
				{\footnotesize (a) cWGAN-gp, $C_L=0.1$. }
			\end{center}
		\end{minipage}%
		\begin{minipage}[h]{0.5\textwidth}
			\begin{center}
				\includegraphics[width=\textwidth]{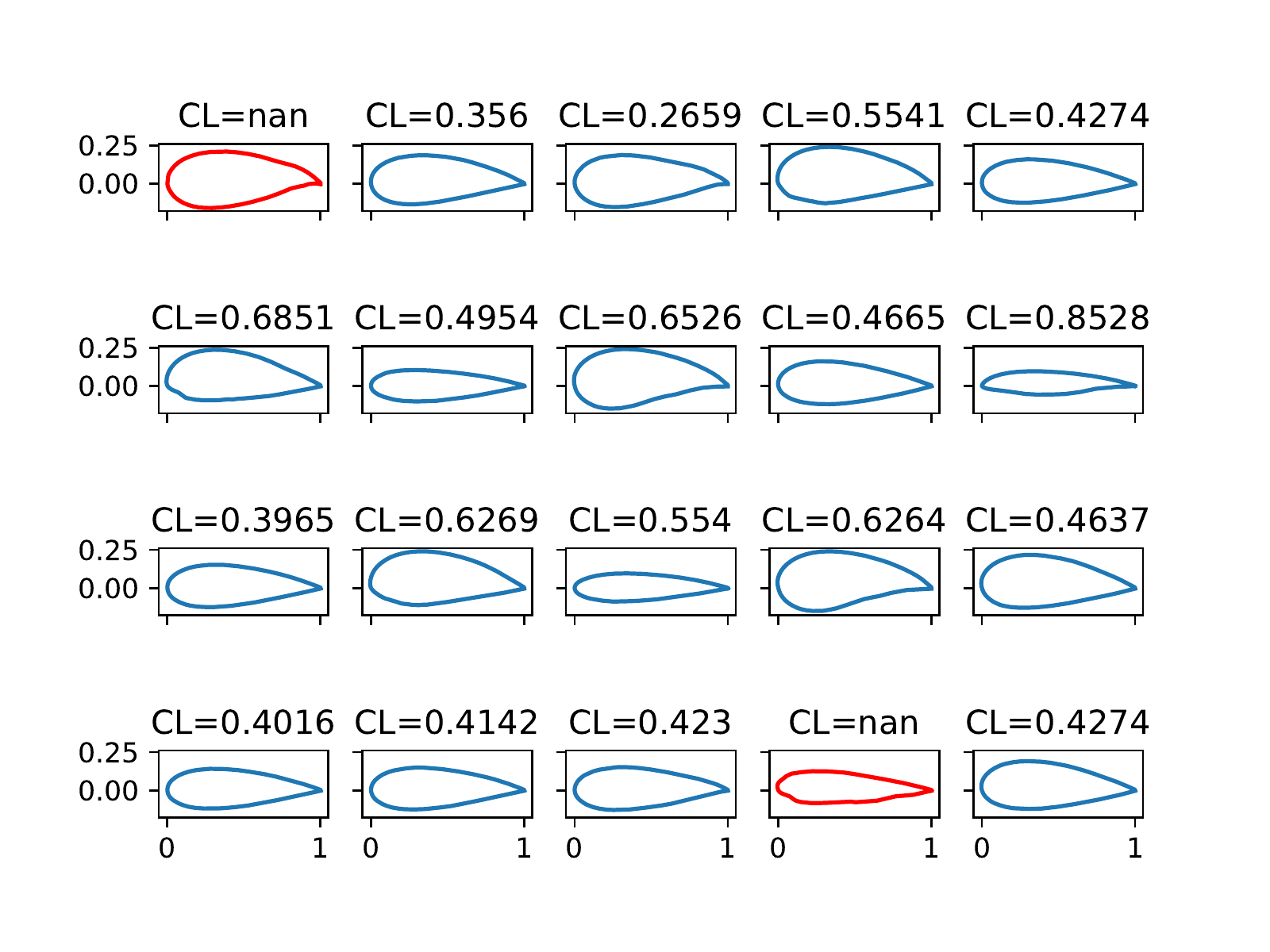}
				\par
				{\footnotesize (b) cWGAN-gp, $C_L=0.5$. }
			\end{center}
		\end{minipage}%
		\par
		\begin{minipage}[h]{0.5\textwidth}
			\begin{center}
				\includegraphics[width=\textwidth]{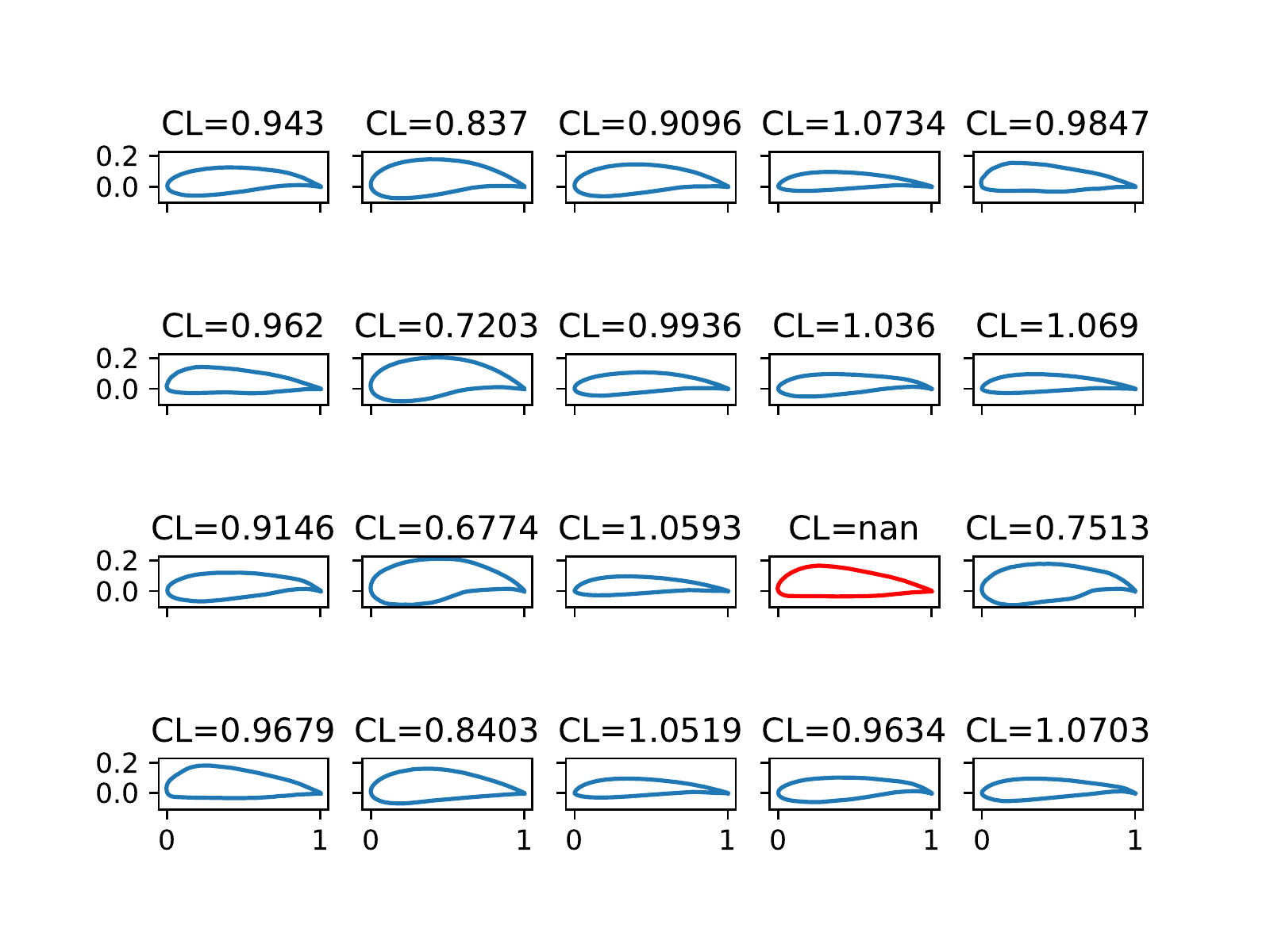}
				\par
				{\footnotesize (c) cWGAN-gp, $C_L=1.0$. }
			\end{center}
		\end{minipage}%
		\begin{minipage}[h]{0.5\textwidth}
			\begin{center}
				\includegraphics[width=\textwidth]{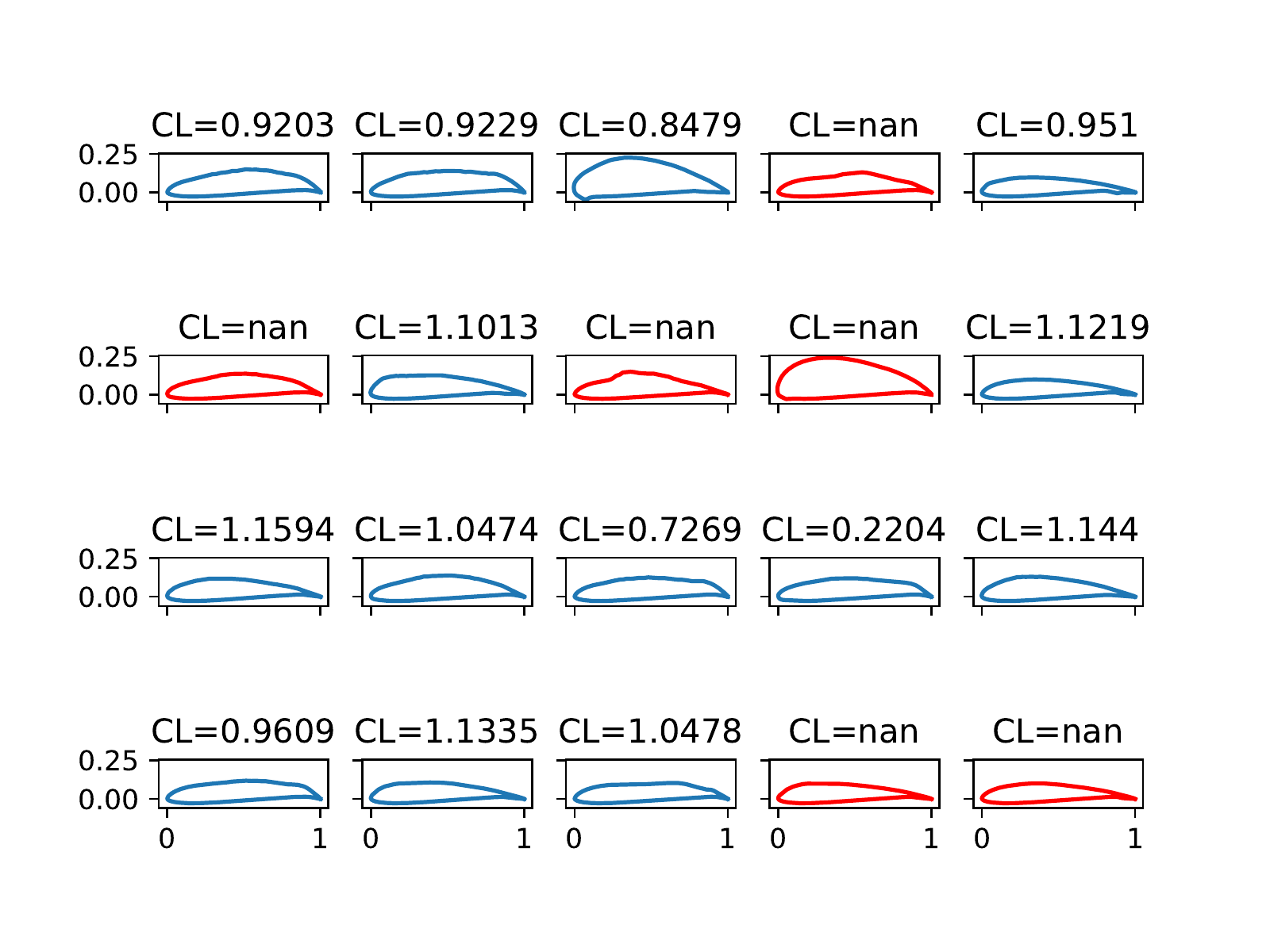}
				\par
				{\footnotesize (d) cWGAN-gp, $C_L=1.4$. }
			\end{center}
		\end{minipage}%
		\caption{Generated shapes of cWGAN-gp ($d=3$). }
		\label{fig:shapes_cWGAN}
	\end{center}		
\end{figure*}

\begin{figure*}[tbph]
	\begin{center}
		\begin{minipage}[h]{\textwidth}
			\begin{center}
				\includegraphics[width=\textwidth]{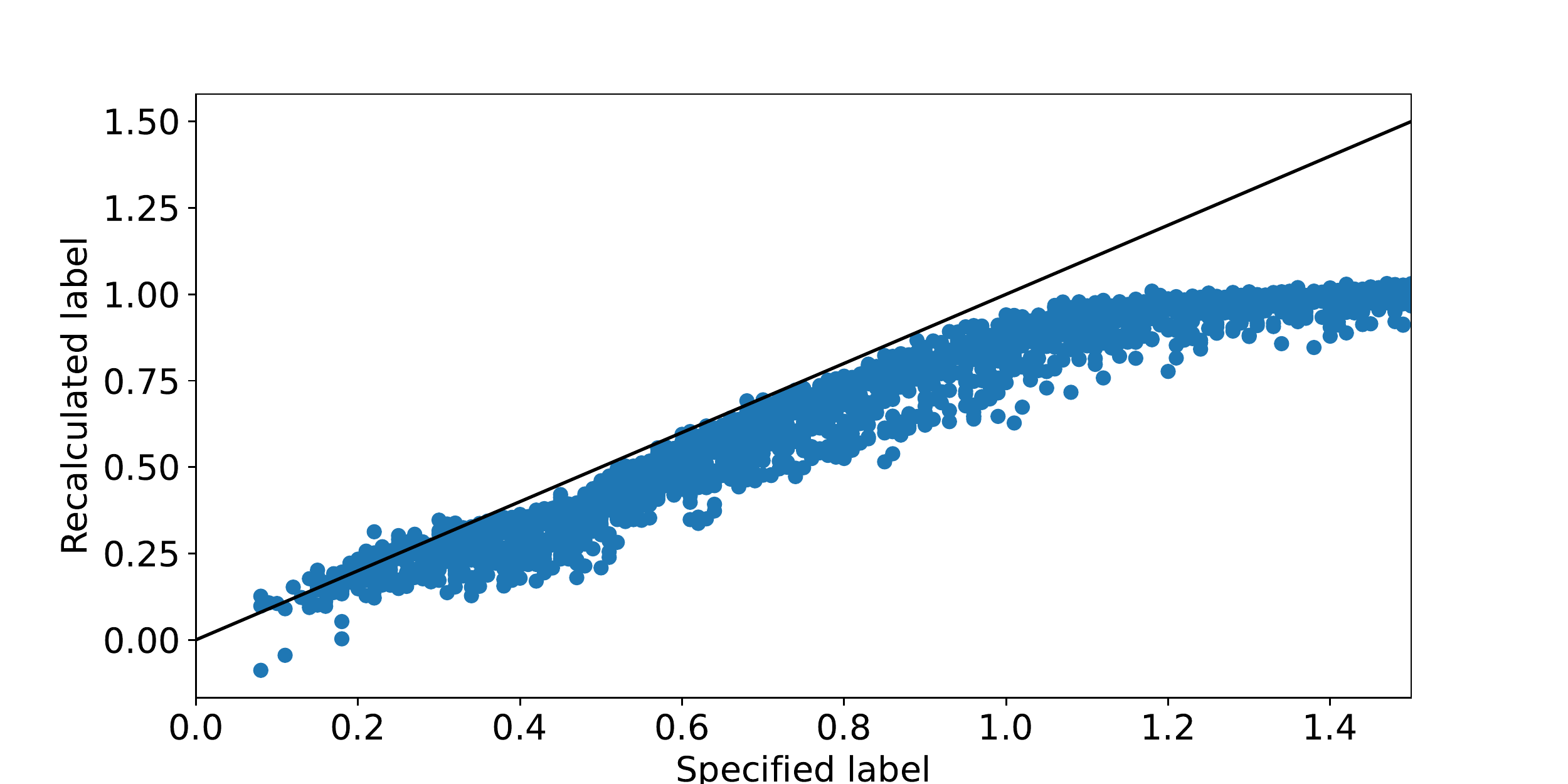}
			\end{center}
			\par
			{\footnotesize (a) Conditional GAN ($d=3$). }
		\end{minipage}%
		\par
		\begin{minipage}[h]{\textwidth}
			\begin{center}
				\includegraphics[width=\textwidth]{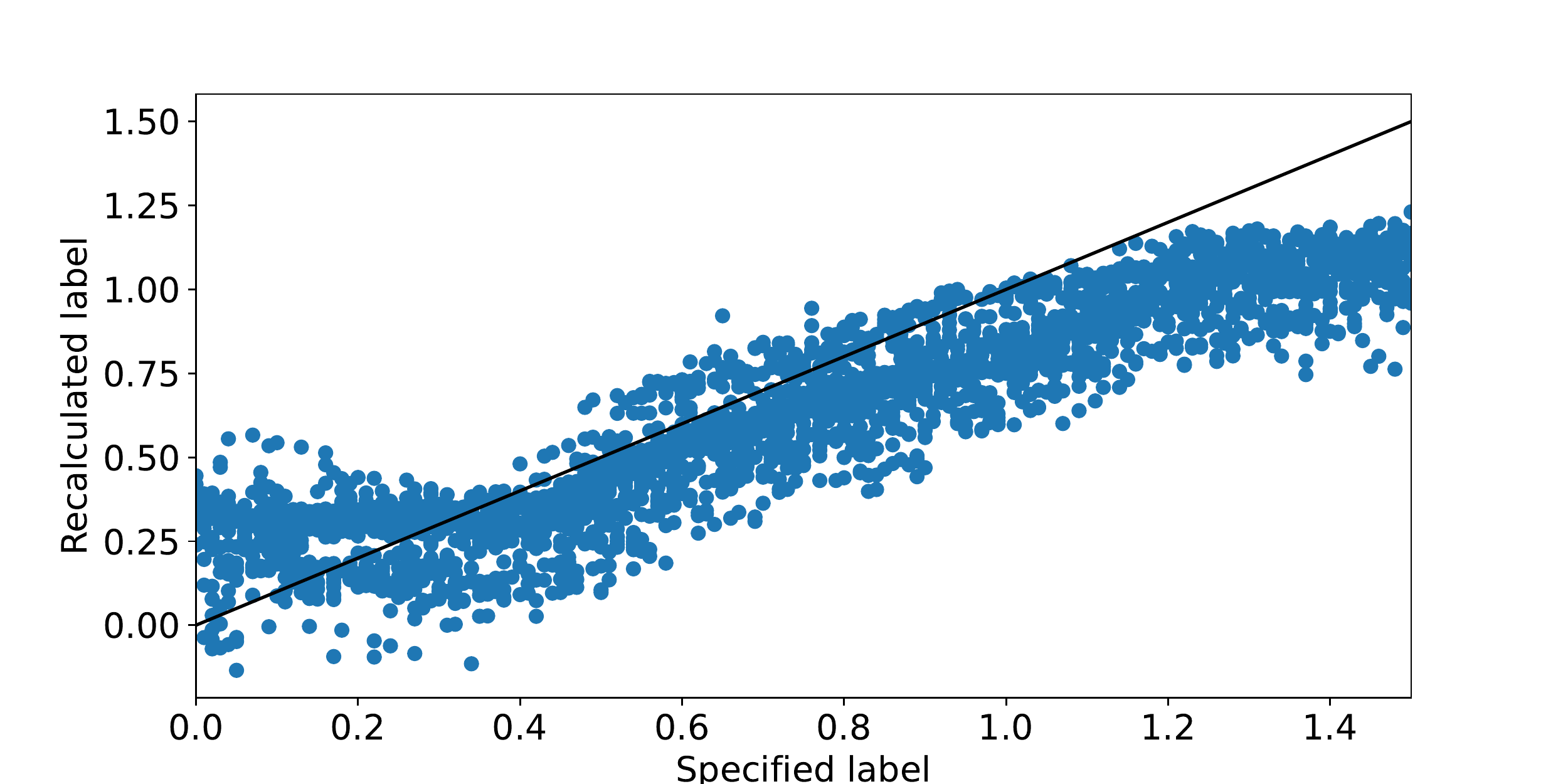}
			\end{center}
			\par
			{\footnotesize (b) Conditional WGAN-gp ($d=3$). }
		\end{minipage}%
		\caption{Error distribution. }
		\label{fig:CL_error}
	\end{center}
\end{figure*}

The labels and recalculated $C_L$s are plotted in Fig. \ref{fig:CL_error} to visualize errors. 
In cGAN, the recalculated $C_L$ was slightly lower in all labels. The highest calculated $C_L$ was approximately $1.0$. 
For the labels lower than 0.2, the points did not appear. It implies that the XFoil calculation of all generated shapes did not converge. 
Conversely, most generated shapes of cWGAN-gp converged. However, the error of the recalculated value and label was higher than that of cGAN. This is attributed to the varieties of generated shapes; the varieties of the shapes of cGAN were less than those of cWGAN. 
Among the generated shapes, those of cGAN and cWGAN at $C_L \in \{ 0,1, 0.5, 1.0, 1.4 \}$ are shown in Fig. \ref{fig:shapes_cGAN} and \ref{fig:shapes_cWGAN}. 
The shapes in blue indicate when the Xfoil computation converged, and those in red indicate otherwise. The recalculated $C_L$s is also shown in the figures. 
In cGAN $C_L=0.1$, almost all the shapes did not converge. The shapes formed airfoil, but they contained small bumps. 
Conversely, in cWGAN-gp $C_L=0.1$, calculations on most shapes converged. 
At $C_L=0.5, 1.0, 1.4$, the XFoil calculations on most shapes obtained from both cGAN and cWGAN-gp converged.

\begin{figure*}[tbph]
	\begin{center}
		\begin{minipage}[h]{\textwidth}
			\begin{center}
				\includegraphics[height=60mm]{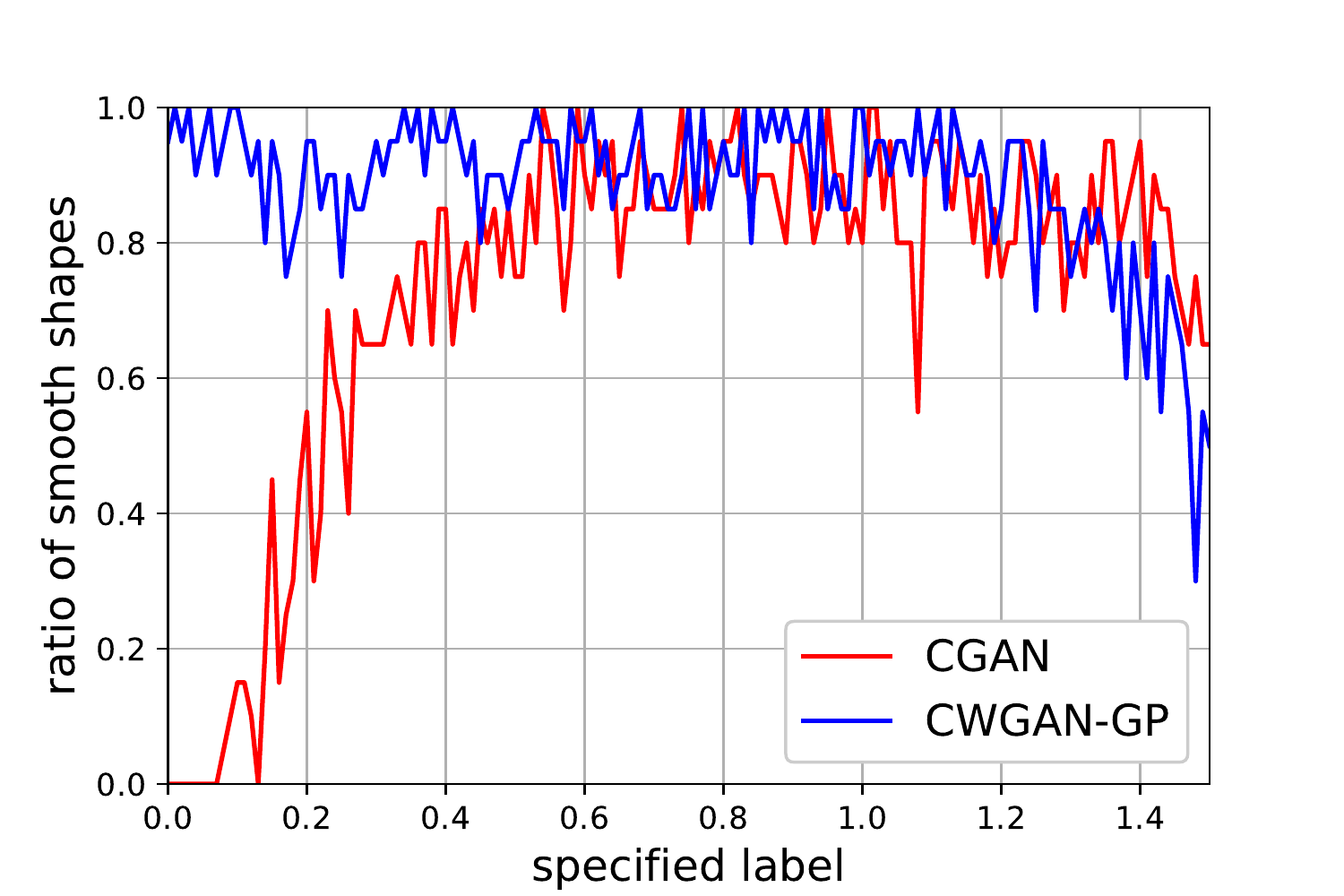}
			\end{center}
			\par
			{\footnotesize (a) Ratio of smooth shapes. }
		\end{minipage}%
		\par
		\begin{minipage}[h]{\textwidth}
			\begin{center}
				\includegraphics[height=60mm]{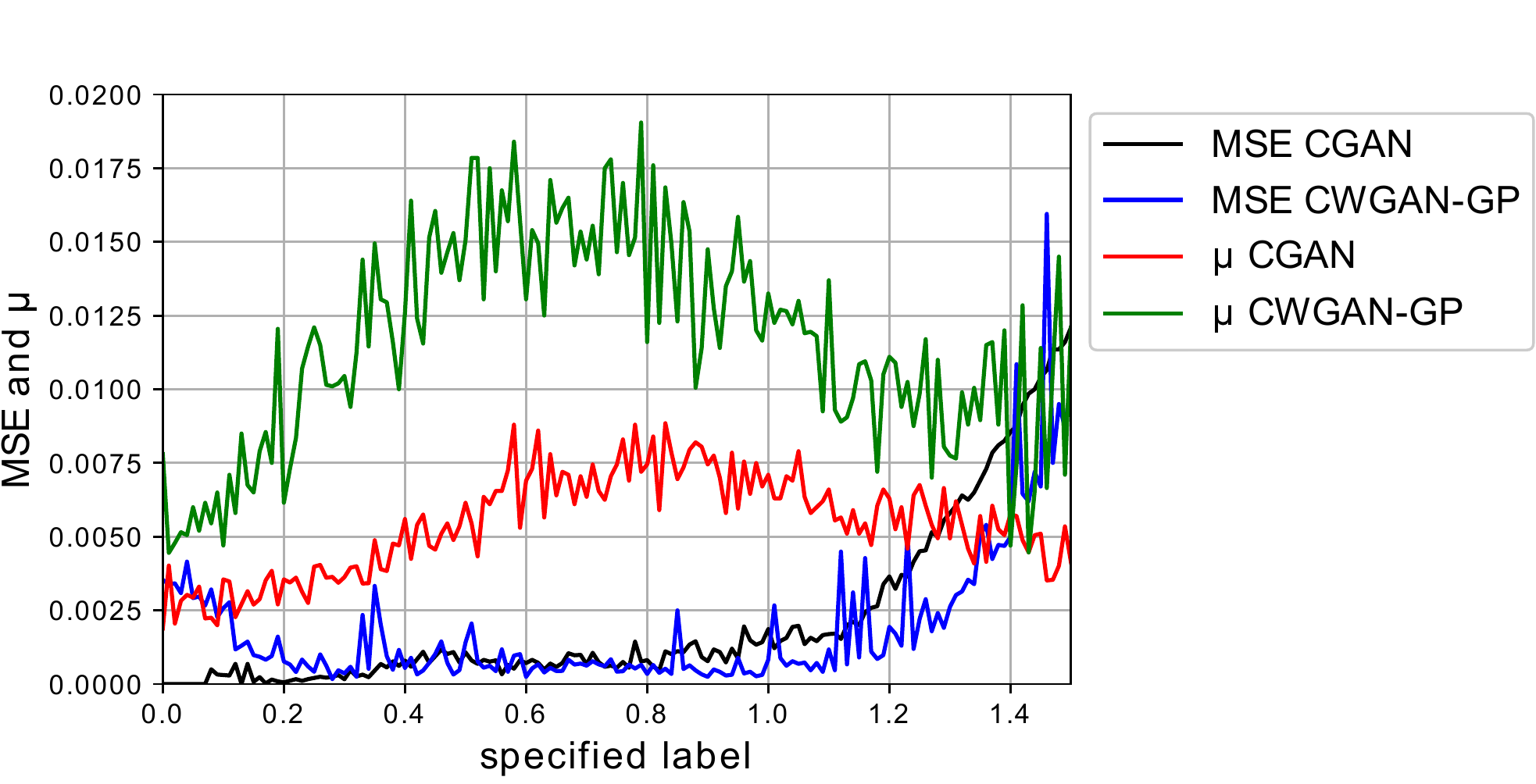}
			\end{center}
			\par
			{\footnotesize (b) MSE and $\mu$. }
		\end{minipage}%
		\caption{Comparison between cGAN and cWGAN-gp. }
		\label{fig:comparison}
	\end{center}
\end{figure*}

The rate of smooth shapes, MSE, and $\mu$ are plotted against the labels in Fig. \ref{fig:comparison}. 
The rate of smooth shapes indicates the shapes whose XFoil computation converges. 
The indices are calculated for each $C_L$. 
The rate of smooth shapes was higher in cWGAN-gp than in cGAN, except at $C_L > 1.3$. 
The MSE of both the models had a similar tendency: it increased as the label increased. 
At $C_L<0.4$, the MSE of cGAN was smaller than that at $C_L>0.4$. In this region, the rate of smooth shapes was small. It implies that there were few smooth shapes among the cGAN outputs, but the shapes satisfied the label with high accuracy. 
The $\mu$ of cWGAN-gp was higher than that of cGAN in all the labels. This is consistent with the fact that WGAN prevents mode collapse.

\section {Conclusion}
Herein, we proposed a cWGAN-gp model for generating airfoil data to overcome the issue of nonsmooth shapes in ordinal GAN models. 
The shapes generated using the cWGAN-gp model were as smooth as those obtained using GAN coupled with a smoothing technique. 
More variation existed in the shapes generated using the cWGAN-gp model than in those using the cGAN model. The difference between cGAN and cWGAN-gp was because WGAN-gp was trained stably, and it avoided mode collapse.
It is beneficial that neither smoothing methods nor free curve representations are required to generate shapes, and it widens the applications of the shape generation method. 

\section{Acknowledgement}
This work was partially supported by JSPS KAKENHI Grant Number 21K14064.

	\bibliographystyle{unsrt}  
	\bibliography{bib-DDD}

\end{document}